%% file: paper.tex
\pdfoutput=1

\documentclass[11pt]{article}

\usepackage[final]{acl}

\usepackage{times}
\usepackage{latexsym}

\usepackage[T1]{fontenc}

\usepackage[utf8]{inputenc}

\usepackage{microtype}
 
\usepackage{inconsolata}

\usepackage{multirow}
\usepackage{booktabs}
\usepackage{graphicx}
\usepackage{amsmath}
\usepackage{xcolor} 
\usepackage{pifont} 
\usepackage{color, colortbl} 
\usepackage{arydshln} 
\usepackage{tabularx} 
\usepackage{footnote} 

\title{\includegraphics[width=0.025\textwidth]{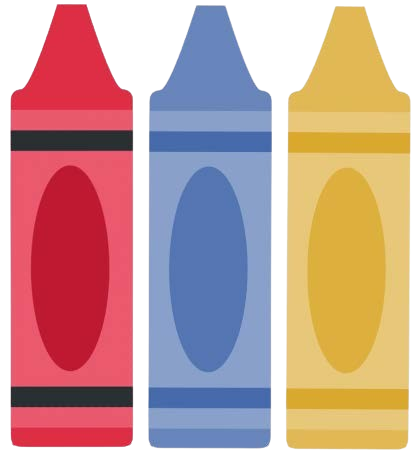} CoLLaVO: Crayon Large Language and Vision mOdel}

\author{Byung-Kwan Lee \\
  KAIST \\
  \texttt{ \small \centering leebk@kaist.ac.kr} \\
  \And
  Beomchan Park \\
  KAIST \\
  \texttt{\small \centering  bpark0810@kaist.ac.kr} \\
  \And 
  Chae Won Kim \\
  KAIST \\
  \texttt{\small \centering  chaewonkim@kaist.ac.kr} \\
  \And
  Yong Man Ro\thanks{Corresponding author.} \\
  KAIST\\
  \texttt{\small \centering  ymro@kaist.ac.kr}
  }

\begin{document}
\maketitle

\begin{abstract} 
The remarkable success of Large Language Models (LLMs) and instruction tuning drives the evolution of Vision Language Models (VLMs) towards a versatile general-purpose model. Yet, it remains unexplored whether current VLMs genuinely possess quality object-level image understanding capabilities determined from `what objects are in the image?' or `which object corresponds to a specified bounding box?'. Our findings reveal that the image understanding capabilities of current VLMs are strongly correlated with their zero-shot performance on vision language (VL) tasks. This suggests that prioritizing basic image understanding is crucial for VLMs to excel at VL tasks. To enhance object-level image understanding, we propose \textbf{C}ray\textbf{o}n \textbf{L}arge \textbf{L}anguage \textbf{a}nd \textbf{V}ision m\textbf{O}del (\includegraphics[width=0.017\textwidth]{figures/crayon_emoji.png} \textbf{CoLLaVO}), which incorporates instruction tuning with \textit{Crayon Prompt} as a new visual prompt tuning scheme based on panoptic color maps. Furthermore, we present a learning strategy of \textit{Dual QLoRA} to preserve object-level image understanding without forgetting it during visual instruction tuning, thereby achieving a significant leap in numerous VL benchmarks in a zero-shot setting. Code is available in \href{https://github.com/ByungKwanLee/CoLLaVO}{https://github.com/ByungKwanLee/CoLLaVO}.
\end{abstract}

\section{Introduction}
\label{sec:introduction}

Spurred by the enduring ambition for artificial general intelligence (AGI) and the success of language models such as BERT~\cite{devlin2018bert}, GPT-3~\cite{brown2020language}, and LLaMA~\cite{touvron2023llama}, there has been a surge in demand for a general-purpose model in a task-unified format via natural language instruction, leading to the emergence of instruction tuning ~\cite{wei2022finetuned, chung2022scaling}. Building on the success of Large Language Models (LLMs) and instruction tuning, InstructBLIP~\cite{dai2023instructblip}, LLaVA1.5~\cite{liu2023visual, liu2023improved}, and Qwen-VL~\cite{bai2023qwen} have either directly designed or utilized visual instruction tuning datasets for a wide range of vision language (VL) tasks using natural language instructions. Consequently, they have become paradigm-shifting in Vision Language Models (VLMs), showcasing remarkable zero-shot performance in VL tasks.

\begin{figure}[t!]
    \vspace{-7mm}
    \centering
    \includegraphics[width=0.48\textwidth]{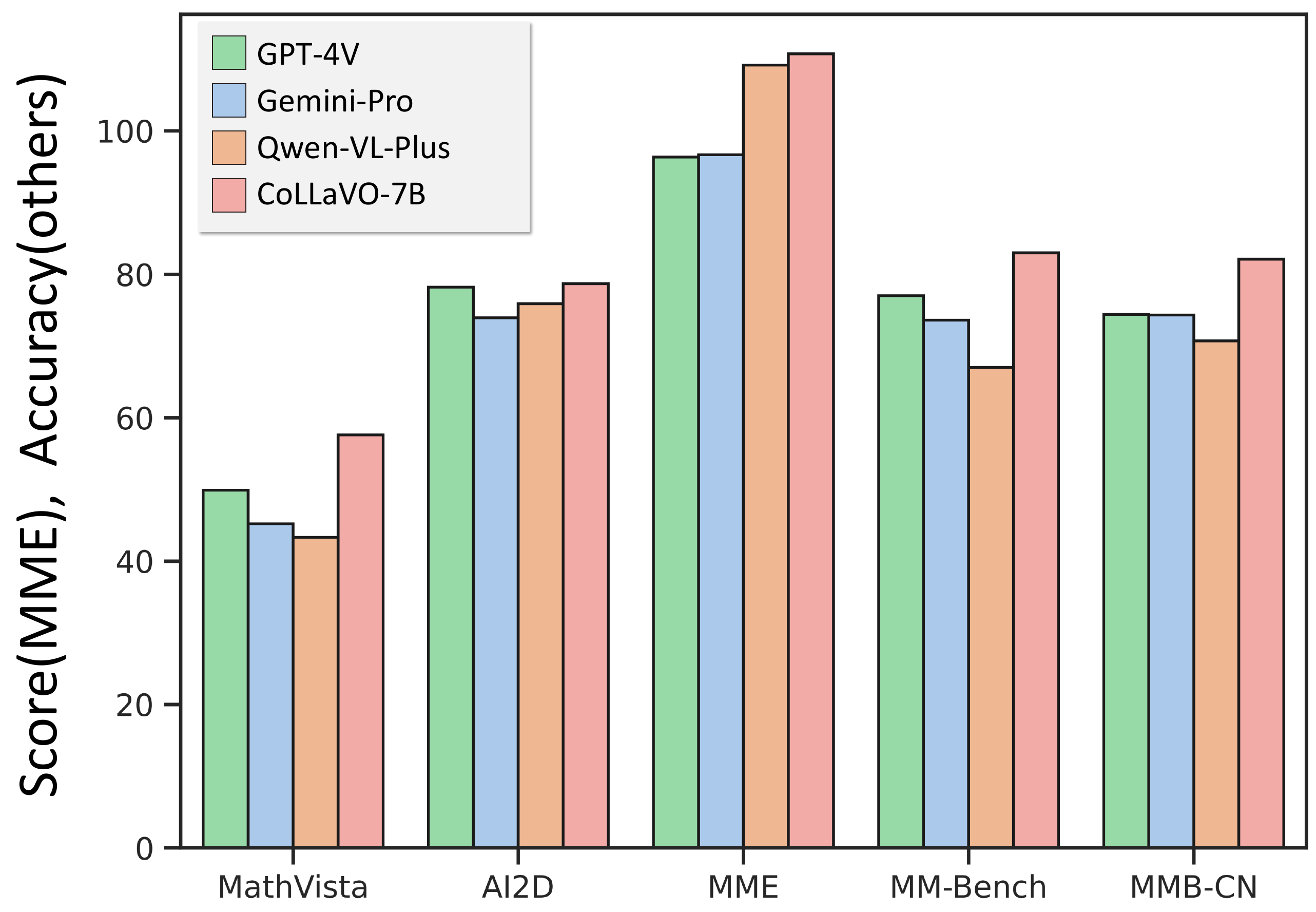}
    \vspace{-6mm}
    \caption{Zero-shot performance of \includegraphics[width=0.017\textwidth]{figures/crayon_emoji.png} CoLLaVO-7B on challenging VL datasets compared with closed-source VLMs~\citep{gptsyscard, gpttechnical,team2023gemini,bai2023qwen}. Note: The scores of MME are rescaled by $1/20$ to match the scales with the accuracies of others.}
    \label{fig:close_vlm}
    \vspace{-5mm}
\end{figure}

\begin{figure*}[t!]
    \vspace{-10mm}
    \centering
    \includegraphics[width=\textwidth]{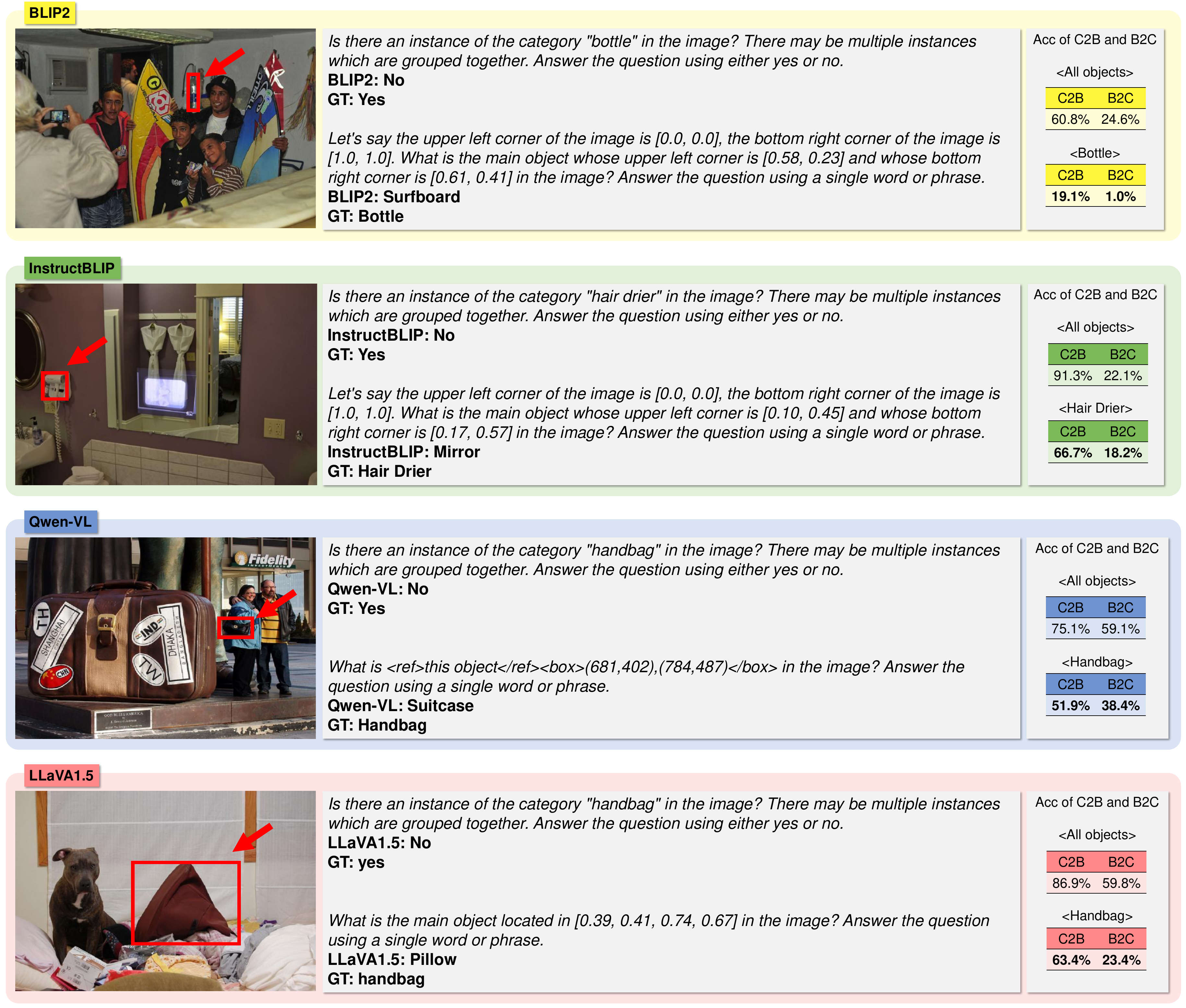}
    \vspace{-6mm}
    \caption{Asking four baselines (BLIP2, InstructBLIP, Qwen-VL, and LLaVA1.5) two types of questions, Class2Binary (C2B) and Box2Class (B2C), and measuring their accuracies on each object category.}
    \label{fig:main_problem}
    \vspace{-2mm}
\end{figure*}

However, it is yet uncharted whether the current leading VLMs truly possess a comprehensive understanding of fine-grained object information, and how this understanding influences their zero-shot performance in VL tasks related to each object. Hence, we delve into the analysis of object-level image understanding and zero-shot performance in VL tasks across different objects. To illustrate the behavior of object-level image understanding, we employ four strong baselines: BLIP2~\citep{blip2}, InstructBLIP~\citep{dai2023instructblip}, LLaVA1.5~\citep{liu2023improved}, and Qwen-VL~\citep{bai2023qwen}. We pose two types of simple questions to gauge their object-level understanding such as: (1) \textit{`Is there any \{object name\} in this image?' (Class2Binary: C2B)}, and (2) \textit{`Which object is in the specified bounding box [$x_{\text{min}}$, $y_{\text{min}}$, $x_{\text{max}}$, $y_{\text{max}}$]?' (Box2Class: B2C)}. We then evaluate the accuracy of their responses for 80 object categories (See Section \ref{par:object-level} for more details) while assessing their zero-shot performance on VL tasks across the same set of categories.

\begin{figure*}[t!]
    \vspace{-5mm}
    \centering
    \includegraphics[width=\textwidth]{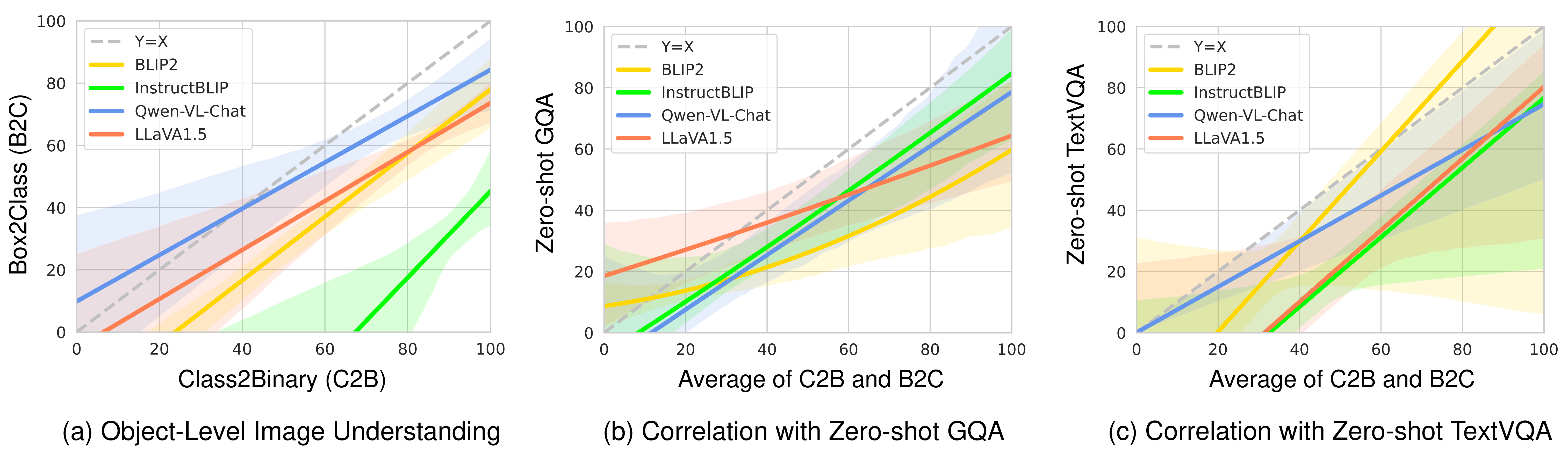}
    \vspace{-5mm}
    \caption{Plotting the regressed relationships between (a) C2B and B2C for each object category, (b) the average of C2B \& B2C and zero-shot GQA~\citep{hudson2019gqa} performance for each object category, (c) the average of C2B \& B2C and zero-shot TextVQA~\citep{singh2019towards} performance for each object category to visualize their correlations. The light-colored areas indicate the vertical span with the probability of confidence interval 0.95.}
    \label{fig:pos_cor}
    \vspace{-4mm}
\end{figure*}

Following this assessment, Figure~\ref{fig:main_problem} illustrates that four strong baselines typically exhibit poor performance on object-level image understanding for several object categories with C2B and B2C accuracies lower than average. This phenomenon arises from various factors, such as biases in co-occurring objects or object size. In Figure~\ref{fig:pos_cor}, we observe a strong correlation between the level of object-level image understanding exhibited by VLMs and their subsequent zero-shot performance. This trend appears consistent across all four baseline VLMs. Consequently, enhancing the object-level image understanding capabilities of VLMs is expected to significantly improve their zero-shot performance in VL tasks.

To improve object-level image understanding, we introduce a new visual prompt called \textit{Crayon Prompt} to assist VLMs in focusing more efficiently on objects. The Crayon Prompt starts from a panoptic segmentation model~\citep{cheng2022masked} that generates a panoptic color map for any given image. This map contains semantic information for objects and their numbering. Leveraging this information, we replace both aspects with learnable queries representing semantic and numbering embeddings, correctly termed as the \textit{Crayon Prompt}.

This simple yet effective idea is inspired by the practice of drawing red circles on images~\citep{shtedritski2023does}, aiming to direct attention to a specific area. They note that red circles potentially invoke the object-level image understanding of VLMs. However, they may distort image contents, posing a risk to VL tasks, and cannot consider foreground and background objects simultaneously. Instead, the Crayon Prompt encompasses all foreground and background objects simultaneously, thanks to a panoptic color map. Unlike drawing a visual prompt directly on an image, we integrate the Crayon Prompt into image embedding features at every attention module layer in the backbone Multi-modal Language Model (MLM) of \includegraphics[width=0.017\textwidth]{figures/crayon_emoji.png} CoLLaVO, thereby keeping the raw visual context of the image intact. The Crayon Prompt imparts semantic information about objects and their numbering, akin to how positional embedding~\citep{vaswani2017attention} assigns sequential information to token embedding features.

By employing the Crayon Prompt, we create simple crayon instructions to enhance object-level image understanding. Additionally, we utilize the visual instruction tuning datasets~\citep{liu2023visual, liu2023improved, chen2023sharegpt4v} for zero-shot VL tasks. However, conducting visual instruction tuning only may be not sure for the grasp of object-level image understanding. Hence, we propose a learning strategy called \textit{Dual QLoRA} involving two QLoRA~\citep{dettmers2023qlora} modules. One module is trained for crayon instructions while the other module for visual instruction tuning datasets is frozen, and vice versa. This approach enables efficient fusion of crayon instructions and visual instruction tuning datasets while preserving the capabilities of both object-level image understanding and complex question answering. Pursuing parameter-efficient training, we employ quantized LoRA (QLoRA) instead of LoRA~\citep{hu2021lora}.

Following the aforementioned methods, we propose a new large language and vision model called \textbf{C}ray\textbf{o}n \textbf{L}arge \textbf{L}anguage \textbf{a}nd \textbf{V}ision m\textbf{O}del (\includegraphics[width=0.017\textwidth]{figures/crayon_emoji.png} \textbf{CoLLaVO}), where the Crayon Prompt and a VLM collaborate to enhance object-level image understanding, which subsequently affects zero-shot VL performance. Our contribution can be summarized as follows:

\begin{itemize}
    \item To the best of our knowledge, we first reveal the intriguing property of current VLMs, wherein object-level image understanding is strongly correlated with zero-shot VL tasks.
    \item We propose the \textit{Crayon Prompt} and \textit{Dual QLoRA}, which enhance object-level image understanding and effectively maintain it alongside complex VL performance, respectively.
    \item By applying all these ingredients, we present an efficient model, \includegraphics[width=0.017\textwidth]{figures/crayon_emoji.png} CoLLaVO-7B, which significantly achieves state-of-the-art zero-shot VL performance compared to closed-source VLMs and open-source VLMs.
\end{itemize}

\section{Research Backgrounds}
\label{sec:related works}

\begin{figure*}[t]
    \vspace{-5mm}
    \centering
    \includegraphics[width=\textwidth]{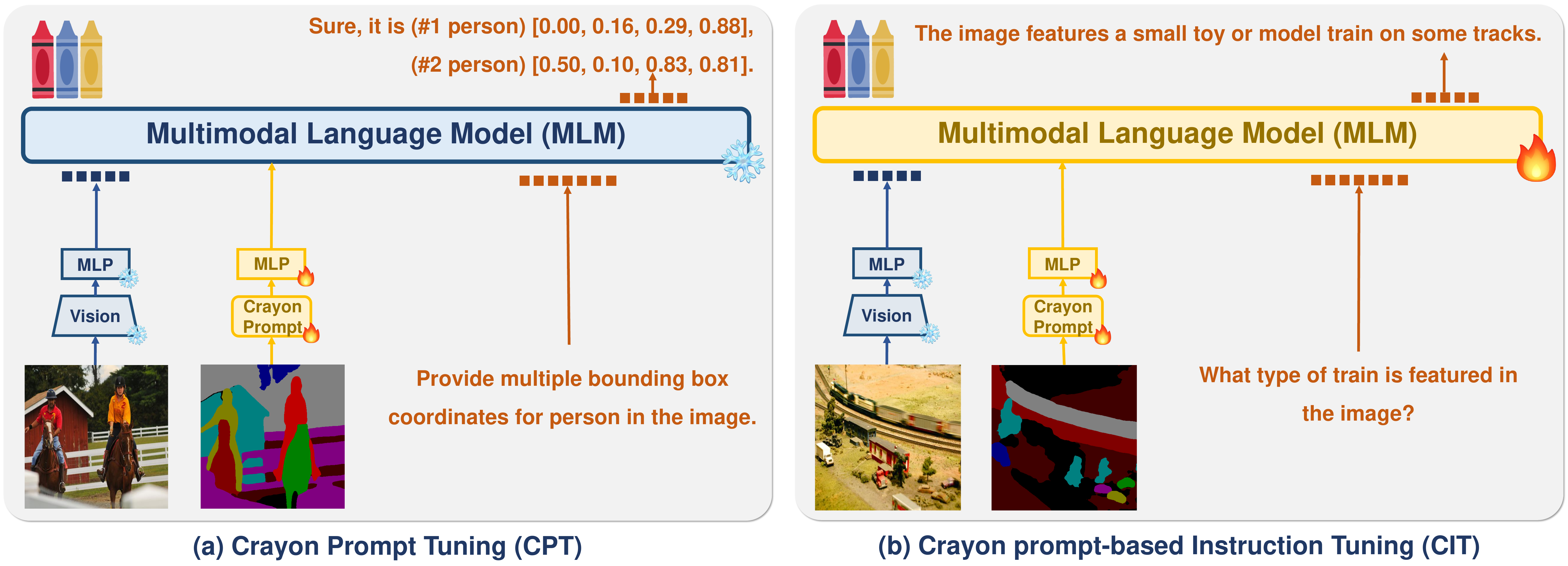}
    \vspace{-5mm}
    \caption{Overview of two-step training for \includegraphics[width=0.017\textwidth]{figures/crayon_emoji.png} CoLLaVO. Note that `Vision' represents vision encoder, and that the fire symbols represent the modules to learn.}
    \label{fig:two_steps}
    \vspace{-4mm}
\end{figure*}

\paragraph{Visual Prompting.} Researchers have prioritized enhancing natural language prompts in constructing instruction tuning datasets for LLMs~\citep{wei2022finetuned, chung2022scaling, touvron2023llama2}. On the other hand, dealing with VLMs offers new opportunities to manipulate both visual and textual aspects of prompts. Earlier studies on visual prompting have focused on techniques such as learnable token embedding concatenated with visual embedding~\citep{jia2022visual, sandler2022fine}, or learned perturbation patterns directly applied to an input image~\citep{bahng2022exploring, chen2023understanding, oh2023blackvip}. While these methods aim to find the optimal visual prompt, the learned visual prompts lack human interpretability, hindering the understanding of their effectiveness.

To address this, current VLMs use human-interpretable visual prompts such as marks~\citep{shtedritski2023does, yang2023finegrained, cai2023making} or semantic masks~\citep{yang2023set}. \citet{shtedritski2023does} draw red circles on images and then demonstrate that CLIP~\citep{clip}, by itself, can recognize the simple visual prompts on images, showing improved zero-shot performance for tasks such as referring expressions comprehension and key point localization. By using SEEM~\citep{zou2023segment} or SAM~\citep{Kirillov_2023_ICCV}, \citet{yang2023set} employs special marks including alphanumerics and masks to help VLMs understand fine-grained spatial information. \citet{yang2023finegrained} uses semantic masks created by an object detection model and SAM, along with visual prompts like contour masks, colorful masks, grayscale reverse masks, and blur reverse masks, to enhance local attention in CLIP.

{
\begin{figure*}[t!]
    \vspace{-10mm}
    \centering
    \includegraphics[width=\textwidth]{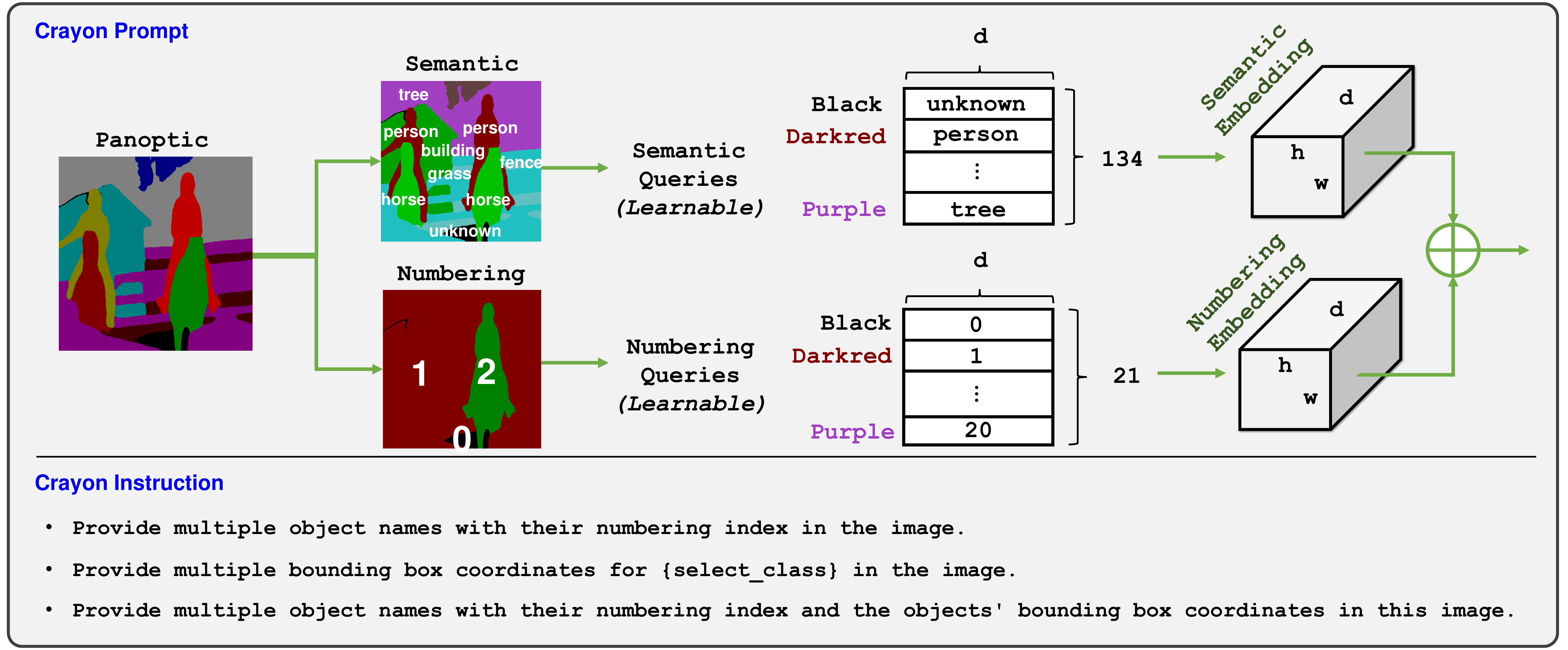}
    \vspace{-5mm}
    \caption{Describing how the Crayon Prompt is generated from a panoptic color map with learnable semantic queries and numbering queries. In addition, crayon instruction examples are given, which are used to conduct CPT and CIT. Note that, `\{\}' denotes the place where we adaptively input information.}
    \label{fig:crayon_prompt}
    \vspace{-3mm}
\end{figure*}
}

In brief, previous studies have focused on guiding VLMs towards specific areas using marks and semantic masks. Similar to \citet{yang2023set}, we propose \textit{Crayon Prompt} encompassing all foreground and background objects at once. However, compared with a direct visual prompt on the image~\citep{liu2023explicit, shtedritski2023does, yang2023finegrained, cai2023making, yang2023set}, the Crayon Prompt is injected into image embedding features at every Transformer~\citep{vaswani2017attention} layer in a backbone MLM to keep the image intact and not disrupt its raw visual context. The Crayon Prompt provides semantic information about objects in the image and their numbering, similar to how positional embedding~\citep{vaswani2017attention} provides sequential information about the relative orders of token embedding features.

\paragraph{LLMs, VLMs, and Instruction Tuning.} Flan~\cite{wei2022finetuned} pioneered the development of instruction tuning by consolidating 62 language datasets, covering a diverse range of tasks. It demonstrates significant improvements in zero-shot performance. In efforts to expand the scope of tasks and the capacity of language models, \citet{chung2022scaling} introduced Flan-PaLM and Flan-T5, leveraging PaLM~\cite{chowdhery2023palm} and T5~\cite{raffel2020exploring}. Continuing along the trajectory of instruction-tuned LLMs, LLaVA~\cite{liu2023visual} utilizes a language-only GPT-4 to produce visual dialogues, intricate deductions, and detailed image descriptions for the LLaVA-Instruct-665K dataset. Simultaneously, various VLMs~\citep{dai2023instructblip,ye2023mplug,li2023otter,zhu2023minigpt,chen2023shikra,bai2023qwen} have developed unique instruction tuning datasets to enhance grounding capability and mitigate hallucinations.

Amidst the current surge of VLMs, we approach them from a fresh angle, notwithstanding the strides made in instruction tuning. Consequently, our focus shifts towards probing whether VLMs effectively grasp object-level image understanding. Should they fall short, we then question whether this inadequacy correlates with their VL performance. In essence, Figure \ref{fig:main_problem}-\ref{fig:pos_cor} emphasize the importance of foundational image understanding and its potential impact on VL performance, in other words, a facet often overlooked in previous studies. Thus, we advocate for a fusion of object-level image understanding and visual instruction tuning.

\section{\includegraphics[width=0.022\textwidth]{figures/crayon_emoji.png} CoLLaVO}
\label{sec:proposed method}

\paragraph{Model Architecture and Prompt Protocol.} The structure of \includegraphics[width=0.017\textwidth]{figures/crayon_emoji.png} CoLLaVO, as illustrated in Figure~\ref{fig:two_steps}, comprises a vision encoder, Crayon Prompt, a backbone MLM, and MLP connectors between the vision and language components. CLIP~\citep{clip} is considered as the vision encoder, benefiting from its adeptness in image understanding. The MLM utilized in \includegraphics[width=0.017\textwidth]{figures/crayon_emoji.png} CoLLaVO is from InternLM-7B~\citep{2023internlm}, which is a multilingual foundation model instruction tuned by 1.6T multilingual datasets with RLHF~\citep{christiano2017deep, stiennon2020learning, ouyang2022training}. Moreover, two fully-connected MLPs with GELU activation function~\citep{hendrycks2016gaussian} serve as the bridge connector. Regarding \includegraphics[width=0.017\textwidth]{figures/crayon_emoji.png} CoLLaVO input, adherence to a prompt protocol is maintained, where `<image>' signifies a special token for image embedding features, `<stop>' denotes a stop token for text generation, `User: \{\}' represents a question template, and `Assistant: \{\}' indicates an answer template (See below Figure~\ref{fig:crayon_prompt} for an example).

\begin{figure*}[t!]
    \vspace{-5mm}
    \centering
    \includegraphics[width=0.8\textwidth]{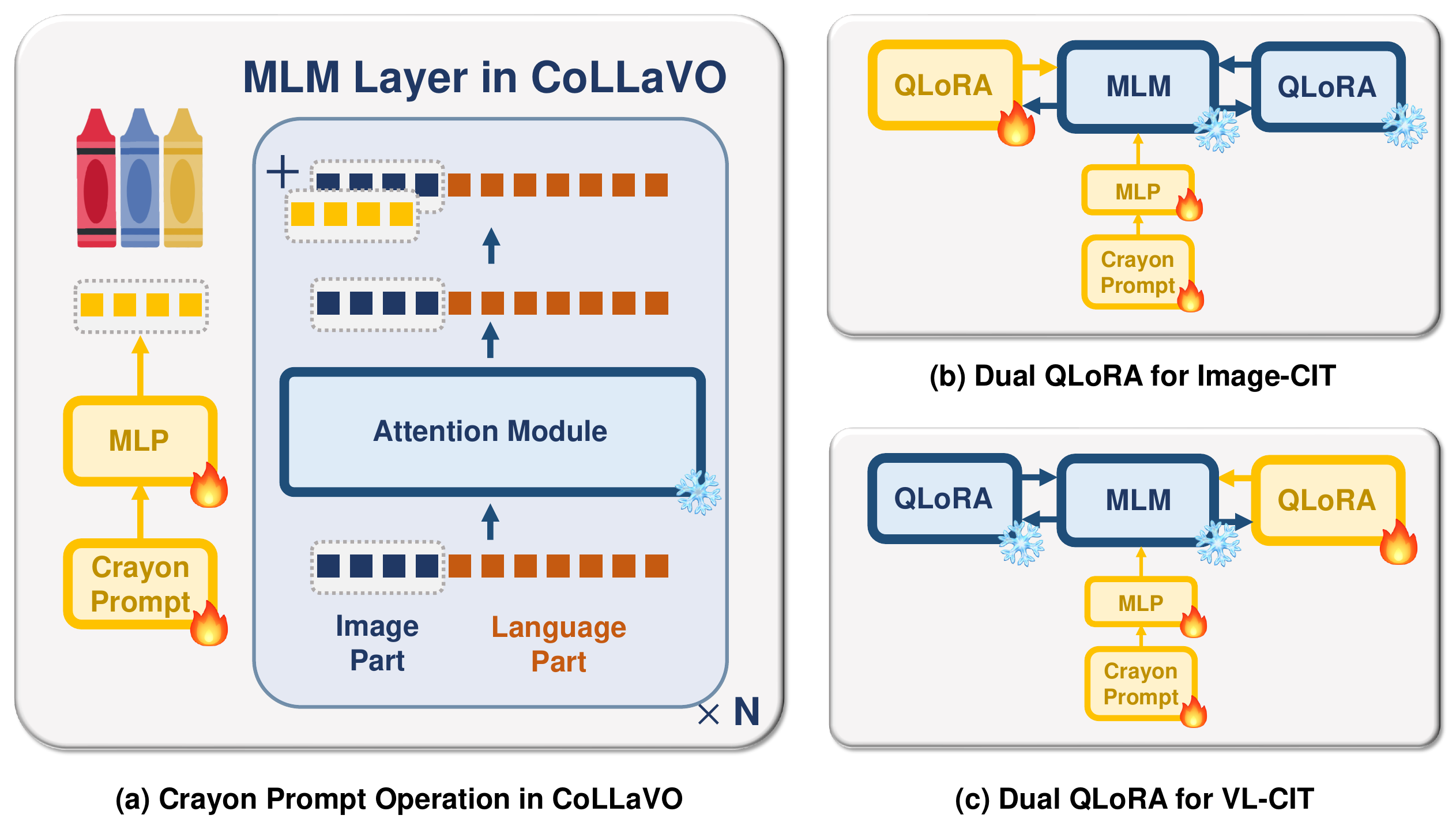}
    \vspace{-3mm}
    \caption{Illuminating (a) how the Crayon Prompt is injected into image embedding features and learning strategies of (b), (c) Dual QLoRA for the object-level image understanding capability (Image-CIT) and VL task capability (VL-CIT) to efficiently coexist without catastrophic forgetting~\citep{luo2023empirical}.}
    \label{fig:dual_qlora}
    \vspace{-3mm}
\end{figure*}

\paragraph{Crayon Prompt Tuning (CPT).} To ensure a comprehensive object-level grasp on the entire image, \includegraphics[width=0.017\textwidth]{figures/crayon_emoji.png} CoLLaVO should recognize all distinct objects within it, including both foreground (\textit{e.g.,} person, bus, bottle, hairdryer, and handbag) and background (\textit{e.g.,} sky, road, river, sea, and snow) objects. To achieve this, we employ a panoptic segmentation model~\citep{cheng2022masked}, which generates a panoptic color map as illustrated in Figure~\ref{fig:two_steps}(a)-(b). This map enables the discrimination of 133 different object categories (See Appendix~\ref{sec:appA}) of foreground and background objects from MS-COCO 2017~\citep{lin2014microsoft}, serving as a visual cue for \includegraphics[width=0.017\textwidth]{figures/crayon_emoji.png} CoLLaVO to focus on all objects within the image. 

Notably, the panoptic map contains two crucial pieces of information for each object: semantic information and numbering information. For instance, if an image depicts two people riding horses, as illustrated in Figure~\ref{fig:two_steps}(a), the panoptic map assigns each object a category label and a numbering index, as shown in Figure~\ref{fig:crayon_prompt}. The two people receive different numbering indices `1' and `2' but share the same object category `person'. Other objects, being singular, are all labeled with the numbering index `1'. It is worth noting that the unknown category is assigned the numbering index `0'. To streamline the next process, we prepare 133+1(\textbf{unk}) learnable semantic queries, including the aforementioned 133 categories and an \textbf{unk}nown category. In addition, we prepare 20+1(`0' for \textbf{unk}) learnable numbering queries under the assumption that no more than 20 instances of the same object category appear within one image. 

\input{tables/table_crayon}

Leveraging 134 semantic queries and 21 numbering queries, we then replace both the semantic and numbering color maps with these queries, akin to generating vector quantized features through a codebook mechanism~\cite{van2017neural, esser2021taming}. This process results in the generation of semantic and numbering embeddings in Figure~\ref{fig:crayon_prompt}, which are subsequently combined in the backbone MLM. This combined representation is referred to as \textit{Crayon Prompt}. The Crayon Prompt meets the MLP connector, and then its output is added with the image features at every attention module layer in the MLM as shown in Figure~\ref{fig:dual_qlora}(a). We then utilize crayon instructions, as shown in the lower half of Figure~\ref{fig:crayon_prompt}, and perform \textit{Crayon Prompt Tuning} (CPT) to align the Crayon Prompt to the backbone MLM and enhance object-level image understanding. Here, \textcolor{magenta}{the magenta colored-text} is auto-regressively learned, as demonstrated in the crayon instruction example below Figure~\ref{fig:crayon_prompt}.

\paragraph {Crayon Prompt-based Instruction Tuning (CIT).} CPT focuses solely on learning semantic and numbering queries in the Crayon Prompt and its MLP connector with the MS-COCO 2017 dataset~\cite{lin2014microsoft}, aligning them with the backbone MLM to enhance object-level image understanding of \includegraphics[width=0.017\textwidth]{figures/crayon_emoji.png} CoLLaVO. On the other hand, \textit{Crayon Prompt-based Instruction Tuning} (CIT) utilizes the visual instruction tuning datasets~\citep{liu2023visual, liu2023improved, chen2023sharegpt4v} as well as crayon instructions to handle complex question answering for VL tasks. It involves training the semantic and numbering queries and the MLP connector again, along with the backbone MLM of \includegraphics[width=0.017\textwidth]{figures/crayon_emoji.png} CoLLaVO. 

When training the MLM with CIT, we introduce a learning strategy called \textit{Dual QLoRA}, which manages object-level image understanding and complex VL performance, respectively, to effectively maintain both aspects. Figure~\ref{fig:dual_qlora} provides an overview of Dual QLoRA, where \textit{Image-CIT} denotes using crayon instructions to bootstrap object-level image understanding and training only the first QLoRA module, while \textit{VL-CIT} indicates using complex question-answer pairs from visual instruction tuning datasets to achieve zero-shot VL performance and training only the second QLoRA module. During CIT, we present an image in the form of Crayon Prompt to \includegraphics[width=0.017\textwidth]{figures/crayon_emoji.png} CoLLaVO, and randomly determine whether to proceed with Image-CIT or VL-CIT. The overarching objective of Dual QLoRA is to efficiently preserve both capabilities of object-level image understanding and complex VL performance. Note that the key distinction between CPT and Image-CIT lies in whether the backbone MLM of \includegraphics[width=0.017\textwidth]{figures/crayon_emoji.png} CoLLaVO is trained or not. Further details will be addressed in the following section.

\begin{figure*}[t!]
    \vspace{-10mm}
    \centering
    \includegraphics[width=\textwidth]{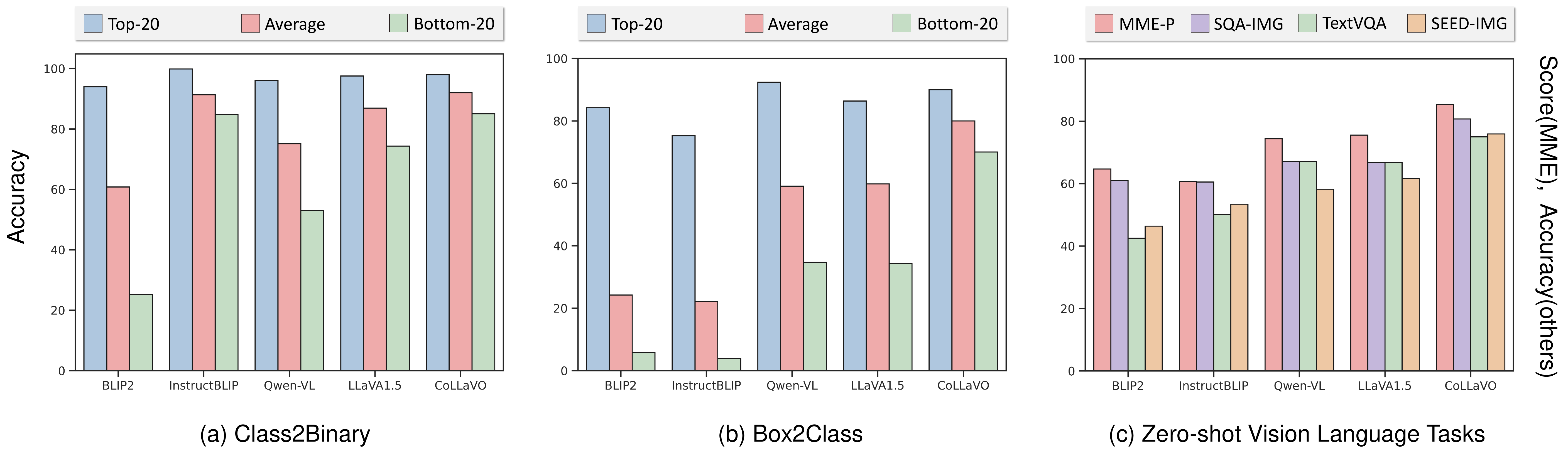}
    \vspace{-6mm}
    \caption{In (a) and (b), there are three metrics for the mean accuracy over Top-20 object categories, Bottom-20, and average of all categories to visualize object-level image understanding of VLMs. In (c), zero-shot performances of VLMs on MME-P (1/20 scaled down of score), SQA-IMG, TextVQA, and SEED-IMG (accuracy) are shown.}
    \label{fig:image_understand}
    \vspace{-4mm}
\end{figure*}

\section{Experiments}
\label{sec:experiments}

\paragraph{Implementation Details of \includegraphics[width=0.017\textwidth]{figures/crayon_emoji.png} CoLLaVO.} To ensure successful reproducibility, we outline the following five crucial technical details of \includegraphics[width=0.017\textwidth]{figures/crayon_emoji.png} CoLLaVO: (a) QLoRA, (b) Crayon Prompt, (c) instruction detail of Image-CIT and VL-CIT, (d) training hyper-parameters, and (e) text-generation.

\textbf{(a)}: we employ Quantized Low-Rank Adaptation (QLoRA)~\citep{hu2021lora, dettmers2023qlora} since \includegraphics[width=0.017\textwidth]{figures/crayon_emoji.png} CoLLaVO pursues efficient training with minimal parameter tuning. Double quantization and normalized float 4-bit (nf4) are used with LoRA of $r=64$ and $\alpha=64$. \textbf{(b)}: In contrast to CPT with only crayon instructions and images from MS-COCO 2017, CIT is conducted with visual instruction tuning datasets~\citep{liu2023visual, liu2023improved, chen2023sharegpt4v} as well. Hence, many images contain unrecognizable objects, such as text, code, posters, or mathematical symbols. Consequently, a panoptic color map with the unknown category and `0' numbering will be generated, and the semantic query of the \textbf{unk} category and numbering query of `0' will operate to create the Crayon Prompt in these cases. \textbf{(c)}: Once the color map is given with discernible objects, text descriptions, including object  names, their numbering indices, and their bounding box coordinates, are added to the question template. Conversely, if an image contains no objects, the question template includes the phrase ``None of detailed object information for image.'' \textbf{(d)}: Regarding training, we train \includegraphics[width=0.017\textwidth]{figures/crayon_emoji.png} CoLLaVO with a batch size of 32 in one epoch using the AdamW~\citep{loshchilov2018decoupled} optimizer, scheduled by cosine annealing~\citep{loshchilov2016sgdr} from a learning rate of 1e-4 to 1e-6 for CPT and from 1e-5 to 1e-6 for CIT, respectively. In addition, $h=35$, $w=35$, and $d=4096$ are used in Figure~\ref{fig:crayon_prompt}. \textbf{(e)}: To find the best performance, \includegraphics[width=0.017\textwidth]{figures/crayon_emoji.png} \textbf{CoLLaVO} uses greedy or beam search ($n=3$) for text generation without any other hyper-parameters.


\input{tables/table1}

\paragraph{Object-level Image Understanding.} 
\label{par:object-level} 
Before delving into validating \includegraphics[width=0.017\textwidth]{figures/crayon_emoji.png} CoLLaVO in VL tasks, it is crucial to ensure its proficiency in object-level image understanding. We assessed the accuracy of 80 object categories classified as `thing' (See Appendix~\ref{sec:appA}) in the MS-COCO 2017 across two directions: Class2Binary (C2B) and Box2Class(B2C), using four strong baselines: BLIP2, InstructBLIP, Qwen-VL, and LLaVA1.5. As illustrated in Figure~\ref{fig:image_understand}(a)-(b), \includegraphics[width=0.017\textwidth]{figures/crayon_emoji.png} CoLLaVO nearly outperforms the baselines in three cases: Top-20, Bottom-20, and Average for both C2B and B2C. Furthermore, it has the smallest performance gap between the Top-20 accuracy and the Bottom-20 accuracy for both C2B and B2C. Such observation indicates that \includegraphics[width=0.017\textwidth]{figures/crayon_emoji.png} CoLLaVO has a solid object-level image understanding across numerous object classes. Beyond its ability, Appendix~\ref{sec:appB} shows zero-shot object grounding performance of \includegraphics[width=0.017\textwidth]{figures/crayon_emoji.png} CoLLaVO for strong generalization to grounding-level understanding.

\input{tables/table2}

\input{tables/table3}

\paragraph{Zero-shot VL Evaluation.} Following improved object-level image understanding, \includegraphics[width=0.017\textwidth]{figures/crayon_emoji.png} CoLLaVO is evaluated to measure zero-shot performance of VL tasks on renowned datasets (See Appendix~\ref{sec:appC}). As shown in Figure~\ref{fig:close_vlm}, \ref{fig:image_understand}(c), and Table~\ref{tab:1}, \includegraphics[width=0.017\textwidth]{figures/crayon_emoji.png} CoLLaVO surpasses several closed-source VLMs like GPT-4V, Gemini-Pro, Qwen-VL-Pro, as well as numerous open-source VLMs (See Appendix~\ref{sec:appD} for all VLMs used in evaluation). Particularly, noteworthy is its superiority over other models in the following benchmarks: MME, MM-Bench, MM-Bench-Chinese, and Q-Bench, which primarily evaluate visual perception and cognition abilities, where \includegraphics[width=0.017\textwidth]{figures/crayon_emoji.png} CoLLaVO demonstrates its significant margins.

\paragraph{The effectiveness of Crayon Prompt and CIT.} We ablate the following factors in \includegraphics[width=0.017\textwidth]{figures/crayon_emoji.png} CoLLaVO: semantic embedding in Crayon Prompt, numbering embedding in Crayon Prompt, Dual QLoRA, Image-CIT, and VL-CIT. As illustrated in Table~\ref{tab:2}, it is evident that the semantic and numbering embedding in the Crayon Prompt significantly boost the zero-shot performance of \includegraphics[width=0.017\textwidth]{figures/crayon_emoji.png} CoLLaVO on MME dataset. It is noteworthy that the semantic embedding alone can improve the zero-shot performance by a large margin, especially in MME-P with `E\&P' scores, implying that injecting object-level semantics helps the model perceive the existence of objects better for solid object-level image understanding. Moreover, the numbering embedding considerably boosts the `Count' score, demonstrating its effectiveness in differentiating objects of the same category by further refining the performance.

Table~\ref{tab:3} demonstrates that Dual QLoRA, Image-CIT, and VL-CIT contribute to improving zero-shot performance, respectively. VL-CIT alone exhibits better performance of $1599.2$ in MME-P and $414.1$ in MME-C over other open-source VLMs, with the assistance of the Crayon Prompt. Additionally, Image-CIT also enhances performance, albeit to a limited extend without QLoRA, by integrating crayon instructions into CIT as well as CPT. Finally, Dual QLoRA produces the most significant improvement, demonstrating its efficacy in fully leveraging both aspects of Image-CIT and VL-CIT.


\begin{figure}[t!]
    \vspace{-3mm}
    \centering
    \includegraphics[width=\linewidth]{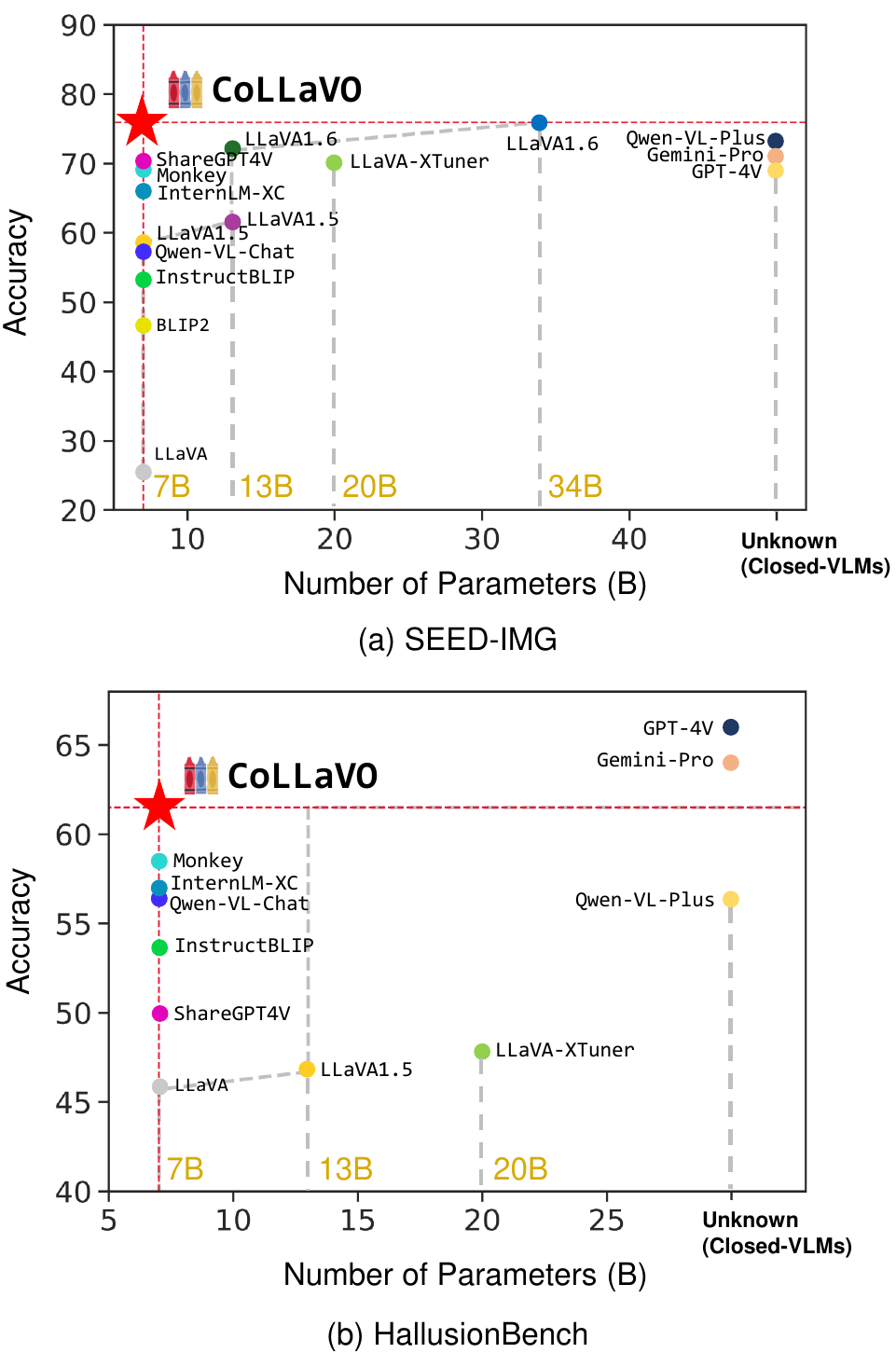}
    \vspace{-5mm}
    \caption{Demonstrating the efficiency and effectiveness of \includegraphics[width=0.017\textwidth]{figures/crayon_emoji.png} CoLLaVO compared with those of other VLMs. Note that accuracy is measured on SEED-IMG and HallusionBench dataset.}
    \label{fig:scale}
    \vspace{-5mm}
\end{figure}

\section{Discussion and Conclusion}
\label{sec:conclusion}
We have shown the effectiveness of \includegraphics[width=0.017\textwidth]{figures/crayon_emoji.png} \textbf{CoLLaVO} alongside \textit{Crayon Prompt} and \textit{Dual QLoRA} serving as a key in enhancing the object-level image understanding. Notably, Figure~\ref{fig:scale}(a) illustrates the impressive ability of \includegraphics[width=0.017\textwidth]{figures/crayon_emoji.png} \textbf{CoLLaVO} achieving cutting-edge zero-shot performance with a relatively small size, thanks to its grasp of object-level understanding validated in SEED-IMG~\citep{li2023seed} with 9 types of questions on spatial understandings of images. Even from the perspective of hallucination, Figure~\ref{fig:scale}(b) and Appendix~\ref{sec:appE} demonstrate that \includegraphics[width=0.017\textwidth]{figures/crayon_emoji.png} \textbf{CoLLaVO} reduces hallucination due to improved object-level image understanding, satisfactorily compared to both closed-source and open-source VLMs on POPE~\citep{li2023evaluating} and HallusionBench~\citep{liu2023hallusionbench}. This suggests that while many researchers have dramatically scaled up their models and curated their own visual instruction tuning datasets, tackling object-level image understanding proves to be an effective strategy.

\section*{Acknowledgments}
This work was partially supported by two funds: IITP grant funded by the Korea government (MSIT) (No.2022-0-00984) and Center for Applied Research in Artificial Intelligence (CARAI) grant funded by DAPA and ADD (UD230017TD).

\section{Limitations}
Crayon Prompts, relying on a panoptic color map, which is an external source beyond VLMs, may be constrained by the performance of the segmentation model and its encompassing number of object classes. Despite this, we have achieved commendable scores across all zero-shot tasks. It is expected for \includegraphics[width=0.017\textwidth]{figures/crayon_emoji.png} \textbf{CoLLaVO} to further improve once it incorporates a plethora of visual prompts obtained from diverse sources like robust object classification or image captioning models~\citep{lee2020towards, lee2022masking, lee2023mitigating, kim2023mitigating}, object-centric causally human-interpretable information~\citep{kim2021distilling, kim2023demystifying}, open object detection~\citep{zhang2023simple}, visual grounding~\citep{liu2023grounding, ren2024grounded}, interactive or unsupervised segmentation~\citep{Kirillov_2023_ICCV, kim2023causal}, optical characteristic recognition model~\citep{bautista2022scene}, and other fascinating approaches~\cite{lee2024meteor, lee2024moai, park2024robust, park2024integrating, kim2024improving}. Beyond its limitation, we believe our promising direction for crayon prompt-like visual cues surely further improve on image understanding for human-like AGI.

\section{Ethics Statement}

We affirm that all research presented in this paper adheres to the principles of ethical conduct and integrity. The experiments conducted and the results reported are based on rigorous scientific methods and strive to contribute positively to the field of vision language models. All datasets used in this study: MS-COCO 2017~\citep{lin2014microsoft} and visual instruction datasets~\citep{liu2023visual, liu2023improved, chen2023sharegpt4v} were obtained and analyzed in compliance with relevant regulations and guidelines for research ethics and data privacy. In addition, any potential limitations have been transparently discussed, so we are committed to upholding the highest standards of integrity, accountability, and respect for communities affected by our research.


\clearpage
\bibliography{reference}

\newpage
\appendix
\onecolumn
\section{COCO Classes for Panoptic Color Map}
\label{sec:appA}

\input{tables/appendixA}

\newpage
\section{Grounding-level Image Understanding}
\label{sec:appB}

\input{tables/table_ref}

\section{Zero-shot Vision Language Datasets used in Evaluation}
\label{sec:appC}

\begin{itemize}

  \item \textbf{GQA}~\citep{hudson2019gqa} is a visual question answering dataset comprising real-world images annotated with scene graphs. It tackles the issue of semantic compositionality by utilizing semantic representations of scenes and questions. It encompasses 22 million questions covering a wide array of images, each associated with structured representations of image objects, attributes, and relations.

  \item \textbf{SQA-IMG}~\citep{iyyer2017search}, a subset of the ScienceQA (SQA) dataset that includes image context, comprises 10,332 multiple-choice questions sourced from elementary and high school science education materials, covering diverse sub-fields. A majority of the questions in the SQA dataset are accompanied by supplementary lectures (83.9\%) and detailed explanations (90.5\%), enriching understanding with broader knowledge and specific reasoning for correct answers.
  
  \item \textbf{TextVQA}~\citep{singh2019towards} is a large-scale complex benchmark to analyze and understand text embedded within images in order to respond to associated questions. This involves integrating textual information present within images and reasoning over it to provide answers. The dataset comprises 28,408 images sourced from OpenImages, accompanied by 45,336 questions and 453,360 corresponding ground truth answers.
  
  \item \textbf{POPE}~\citep{li2023evaluating} serves as a polling-based binary classification query dataset, tailored to assess object hallucination challenges within VLMs. It comprises three distinct subsets, \textit{i.e.,} random, popular, and adversarial, each crafted using varied sampling techniques, resulting in a total of 8,910 entries.

  \item \textbf{MME}~\citep{fu2023mme} is introduced as a novel comprehensive benchmark aimed at assessing the performance of VLMs by measuring both perception and cognition abilities across 14 sub-tasks. To mitigate potential data leakage issues associated with public datasets, all annotations for instruction-answer pairs are manually designed.
  
  \item \textbf{MMBench, MMBench-Chinese}~\citep{liu2023mmbench} establish a comprehensive evaluation framework spanning multiple modalities. These frameworks encompass around 3000 multiple-choice questions addressing 20 distinct capability dimensions in both English and Chinese languages. An innovative approach is introduced through the integration of ChatGPT into the evaluation process.
  
  \item \textbf{MM-Vet}~\citep{yu2023mm} is a multi-modal assessment benchmark that assesses a broad range of capabilities essential for handling real-world scenarios, such as solving mathematical problems or interpreting visual humor. The dataset consists of 187 images collected from diverse online platforms and presents 205 questions, each requiring the application of one or more capabilities for an answer. These questions vary in type and necessitate open-ended responses of varying lengths.
  
  \item \textbf{Q-Bench}~\citep{wu2023q} evaluates VLMs across three dimensions relevant to low-level vision: perception, description, and assessment. To assess perception, the framework utilizes 2,990 diverse images, each accompanied by a human-generated question focusing on its low-level attributes. For evaluating VLMs' description regarding low-level information, human-labeled textual descriptions for 499 images are utilized, alongside a comparison pipeline involving GPT. Additionally, the framework assesses VLMs' visual quality assessment abilities, aiming to align with human opinion scores.
  
  \item \textbf{MathVista}~\citep{lu2023mathvista} assesses VLMs' mathematical reasoning ability within visual contexts, with 6,141 examples sourced from 28 existing multimodal datasets on mathematics. MathVista provides a comprehensive evaluation platform, requiring meticulous visual comprehension and compositional reasoning, posing challenges even to state-of-the-art foundational models.
  
  \item \textbf{AI2D}~\citep{kembhavi2016diagram}, or AI2 Diagrams, is a dataset comprising over 5,000 grade school science diagrams. It includes comprehensive annotations of constituents and relationships, along with rich syntactic parses and over 15,000 corresponding multiple-choice questions.

  \item \textbf{SEED-IMG}~\citep{li2023seed} comprises a subset of SEED-Bench, focusing on the image modality. The original SEED-Bench includes 19,000 multiple-choice questions with precise human annotations, covering 12 evaluation dimensions, including comprehension of both image and video modalities.
  
  \item \textbf{HallusionBench}~\citep{liu2023hallusionbench} introduces a comprehensive benchmark tailored for evaluating image-context reasoning abilities. It prioritizes nuanced comprehension and interpretation of visual information. The benchmark consists of 346 images accompanied by 1129 expert-crafted questions, enabling a quantitative analysis of model response tendencies, logical consistency, and diverse failure modes.

\end{itemize}

\section{Vision Language Models used in Evaluation}
\label{sec:appD}

\begin{itemize}
  \item \textbf{BLIP2}~\citep{blip2} introduces Q-Former that serves as an intermediary between frozen unimodal models, extracting pertinent visual features from a frozen image encoder and providing them to a frozen large language model to generate text.
  
  \item \textbf{InstructBLIP}~\citep{dai2023instructblip} presents a vision-language instruction tuning framework designed to address the challenges of generalizing to diverse tasks, through a systematic study involving 26 datasets transformed into instruction tuning format across 11 task categories.

  \item \textbf{Shikra}~\citep{chen2023shikra} proposes a unified model designed for referential dialogue tasks, which encompass various vision-language tasks such as VQA, image captioning, and location-related tasks like referring expression comprehension and PointQA.

  \item \textbf{IDEFICS}~\citep{laurenccon2023obelisc} introduces a curated web-scale dataset comprising 141 million multimodal English web documents, each containing associated images and text, totaling 353M images and 115B tokens. They aim to provide full multimodal documents preserving the natural context of images within web pages.

  \item \textbf{Qwen-VL, Qwen-VL-Chat}~\citep{bai2023qwen} introduces Qwen-VL series, a collection of highly performant and versatile vision-language models based on Qwen language model. They support multiple languages and handling of multi-image inputs, and fine-grained visual understanding capabilities.

  \item \textbf{MiniGPT-4}~\citep{zhu2023minigpt} presents a vision-language model that combines Vicuna with freezed pre-trained vision components of Q-Former from BLIP2, aiming to replicate the exceptional capabilites demonstrated by GPT-4.

  \item \textbf{MiniGPT-v2}~\citep{chen2023minigpt} is designed to effectively handle multiple vision-language tasks by employing a task-oriented instructiom training scheme, through three training stage and utilization of higher-resolution images.

  \item \textbf{Otter}~\citep{li2023otter} addresses the gap between DeepMind Flamingo by employing OpenFlamingo and multi-modal in-context instruction tuning (MIMIC-IT) dataset.

  \item \textbf{LLaVA}~\citep{liu2023visual, liu2023improved} first introduces the concept of visual instruction tuning, extending language only instruction tuning to vision language instruction tuning to develop a general-purpose visual assistant.   
  
  \item \textbf{LLaVA-XTuner}~\citep{2023xtuner} is a tool to fine-tune LLaVA to achieve general-purpose model.

  \item \textbf{mPLUG-Owl}~\citep{ye2023mplug} introduces a modularized training paradigm for large multi-modal language models capable of supporting multiple modalities simultaneously. Inspired by modularization concepts, their method integrates pre-trained language models, visual knowledge modules, and visual abstractor modules to achieve effective alignment between images and text.

  \item \textbf{mPLUG-Owl2}~\citep{ye2023mplug2} features a modularized network design to handle both modality collaboration and interference. They introduce shared functional modules to promote collaboration and a modality-adaptive module to manage different modalities effectively.

  \item \textbf{ShareGPT4V}~\citep{chen2023sharegpt4v} argues that current Large multi-modal models face sub-optimal modality alignment due to the lack of high-quality image-text pairs. To address this issue, they collected high-quality captions on a larger scale in two phases. This effort led to the creation of the ShareGPT4V dataset, comprising 100K GPT4-Vision generated captions and 1.2M captions crafted by their caption model.

  \item \textbf{CogVLM}~\citep{wang2023cogvlm} handles challenges of the lack of direct equivalence between visual and textual input spaces. They introduce a trainable visual expert to the language model, where it allows for the retention of natural language processing capabilities while enhancing visual understanding abilities.

  \item \textbf{Intern-XC}~\citep{zhang2023internlm} is trained to generate long-form content interleaved with contextually relevant images, based on a multilingual vision-language dataset comprising over 11M semantic concepts collected from public websites, thereby enhancing vision-language interactions.
  
  \item \textbf{MM-GPT}~\citep{gong2023multimodal} fine-tune OpenFlamingo using comprehensive datasets of image and text instructions to conduct multi-turn image-text dialogues more closely aligned with human preferences. A perceiver resampler is used for efficient visual information extraction and gated cross-attention layers for image-text interactions.
\end{itemize}

\newpage
\section{Detail of POPE dataset for Hallucination}
\label{sec:appE}
\input{tables/appendixD}


\end{document}

%% file: tables/table_crayon.tex
\begin{table}[t!]
\resizebox{\linewidth}{!}{
\renewcommand{\tabcolsep}{0.01mm}
\centering
\begin{tabular}{l}
\toprule
Crayon Instruction Example for CPT
\\
\midrule
<image>\\
User: Provide multiple object names with their numbering index\\
and the objects' bounding box coordinates in this image.\\
Assistant: \textcolor{magenta}{Sure, it is (\#1 horse) [0.06, 0.38, 0.27, 0.91],} \\
\textcolor{magenta}{(\#2 horse) [0.56, 0.41, 0.77 , 0.99], (\#1 person) [0.00, 0.16, 0.29, 0.88],} \\
\textcolor{magenta}{(\#1 sky) [0.39, 0.00, 0.69, 0.18], (\#1 fence) [0.00, 0.42, 0.99, 0.99],} \\
\textcolor{magenta}{(\#1 building) [0.00, 0.14, 0.47, 0.83], (\#1 tree) [0.00, 0.00, 0.99, 0.47],} \\
\textcolor{magenta}{(\#1 grass) [0.21, 0.49, 0.99, 0.95], (\#2 person) [0.50, 0.10, 0.83, 0.81]<stop>} \\
\bottomrule
\end{tabular}
}
\label{prompt_example}
\vspace{-5mm}
\end{table}

%% file: tables/table1.tex
\definecolor{Gray}{gray}{0.93}
\definecolor{Green}{rgb}{0.9, 0.95, 0.97}

\begin{table*}[t!]
\vspace{-7mm}
\centering
\resizebox{\linewidth}{!}{
\renewcommand{\tabcolsep}{2mm}
\begin{tabular}{lcccccccccccc}
\toprule
VLM             & GQA  & SQA-IMG   & TextVQA & POPE & MME-P & MME-C & MM-Bench & MMB-CN         & MM-Vet & Q-Bench \\
\midrule
\rowcolor{Gray}
BLIP2-13B       & 42.4 & 61.0      & 42.5    & 85.3 & 1293.8& 290.0 & -        & -              & 22.4   & -       \\
\rowcolor{Gray}
InstructBLIP-7B & 49.2 & 60.5      & 50.1    & -    & -     & -     & 36.0       & 23.7           & 26.2   & 56.7    \\
\rowcolor{Gray}
InstructBLIP-13B& 49.5 & 63.1      & 50.7    & 78.9 & 1212.8& -     & -        & -              & 25.6   & -       \\
Shikra-13B      & -    & -         & -       & -    & -     & -     & 58.8     & -              & -      & 54.7    \\
IDEFICS-9B      & 38.4 & -         & 25.9    & -    & -     & -     & 48.2     & 25.2           & -      & -       \\
IDEFICS-80B     & 45.2 & -         & 30.9    & -    & -     & -     & 54.5     & 38.1           & -      & -       \\
\rowcolor{Gray}
Qwen-VL-7B      & 59.3 & 67.1      & 63.8    & -    & -     & -     & 38.2     & 7.4            & -      & 59.4    \\
\rowcolor{Gray}
Qwen-VL-Chat-7B & 57.5 & 68.2      & 61.5    & -    & 1487.5& 360.7 & 60.6     & 56.7           & -      & -       \\
MiniGPT-4-7B    & 43.5 & -         & -       & -    & 581.7 & -     & 23.0     & -              & 22.1   & -       \\
Otter-7B        & -    & -         & -       & -    & 1292.3& -     & 48.3     & -              & 24.6   & 47.2    \\
LLaVA-7B        & -    & 38.5      & -       & -    & 807.0 & 247.9 & 34.1     & 14.1           & 26.7   & -       \\
MiniGPT-v2-7B   & 60.3 & -         & -       & -    & -     & -     & -        & -              & -      & -       \\
MiniGPT-v2-Chat-7B & 60.1 & -         & -       & -    & -     & -     & -        & -              & -      & -       \\
\rowcolor{Gray}
LLaVA1.5-7B     & 62.0 & 66.8      & 58.2    & {85.9} & 1510.7& 293.8& 64.3     & 58.3           & 30.5   & 58.7    \\
\rowcolor{Gray}
LLaVA1.5-13B    & \textbf{63.3} & {71.6}      & {61.3}    & {85.9} & 1531.3& 295.4 & 67.7     & 63.6           & 35.4   & 62.1    \\
mPLUG-Owl-7B    & -    & -         & -       & -    & 967.3 & -     & 46.6     & -              & -      & 58.9    \\
mPLUG-Owl2-7B   & 56.1 & 68.7      & 58.2    &      & 1450.2& -     & 64.5     & -              & 36.2   & 62.9    \\
ShareGPT4V-7B   & -  & 68.4      & -       &      & {1567.4}& 376.4 & 68.8     & 62.2              & {37.6}   & 63.4    \\
CogVLM-17B      & 56.1 & 68.7      & 58.2    &      & -     & -     & 65.8     & 55.9           & \textbf{54.5}   & -       \\
LLaVA-XTuner-20B& -    & -         & -       & -    & -     & -     & {75.1}     & {73.7}           & 37.2   & -       \\
Intern-XC-7B    & -    & -         & -       &      & 1528.4& {391.1} & 74.4     & 72.4           & 35.2   & {64.4}    \\
\midrule
\rowcolor{Green} 
CoLLaVO-7B      & 61.4  & \textbf{80.7}      & \textbf{64.2}    & \textbf{87.2} & \textbf{1689.7}& \textbf{525.0}& \textbf{83.0}    & \textbf{82.1}           & 40.3   & \textbf{67.6}    \\
\bottomrule 
\end{tabular}
}
\vspace{-2mm}
\caption{Evaluating zero-shot performances of \includegraphics[width=0.017\textwidth]{figures/crayon_emoji.png} CoLLaVO on ten vision language datasets compared with the current powerful VLMs such as InstructBLIP, Qwen-VL, LLaVA1.5, and so forth.}
\label{tab:1}
\vspace{-3mm}
\end{table*}

%% file: tables/table2.tex
\newcommand{\cmark}{\ding{51}}%
\newcommand{\xmark}{\ding{55}}%
\begin{table}[t!]
\centering
\resizebox{\linewidth}{!}{
\renewcommand{\tabcolsep}{2.0mm}
\begin{tabular}{cccccc}
\toprule
\multicolumn{2}{c}{Crayon Prompt}  & \multicolumn{4}{c}{MME} \\
\cmidrule(lr){1-2}\cmidrule(lr){3-6}
Sem-Query & Num-Query & MME-P& MME-C & E\&P  & Count      \\
\midrule
\xmark  & \xmark  & 1553.4   & 375.0 & 288.3 & 141      \\
\cdashline{1-6}\noalign{\vskip 0.5ex}
\cmark  & \xmark  & 1636.7   & 482.1 & 310.6 & 147      \\
\cdashline{1-6}\noalign{\vskip 0.5ex}
\rowcolor{Green}
\cmark  & \cmark  & \textbf{1689.7}   & \textbf{525.0} & \textbf{341.6} & \textbf{160}       \\
\bottomrule
\end{tabular}
}
\caption{Controlling semantic and numbering queries in crayon prompt. Note: `E\&P' denotes the score of the existence and position, and `Count' denotes the score to understand the numbering.}
\vspace{-3mm}
\label{tab:2}
\end{table}

%% file: tables/table3.tex
\begin{table}[t!]
\centering
\resizebox{\linewidth}{!}{
\renewcommand{\tabcolsep}{0.6mm}
\begin{tabular}{ccccccc}
\toprule
\multicolumn{3}{c}{CIT}  & \multicolumn{4}{c}{MME} \\
\cmidrule(lr){1-3}\cmidrule(lr){4-7}
Dual Q-LoRA & Image-CIT & VL-CIT & MME-P    & MME-C  & E\&P  & Count     \\
\midrule
\xmark & \xmark  & \cmark  & 1599.2   & 414.1 & 298.6 & 145     \\
\cdashline{1-7}\noalign{\vskip 0.5ex}
\xmark & \cmark  & \cmark  & 1620.5   & 456.2 & 308.3 & 146     \\
\cdashline{1-7}\noalign{\vskip 0.5ex}
\rowcolor{Green}
\cmark & \cmark  & \cmark  & \textbf{1689.7}   & \textbf{525.0} & \textbf{341.6} & \textbf{160}  \\
\bottomrule
\end{tabular}
}
\caption{Controlling Dual QLoRA, Image-CIT, and VL-CIT in conducting CIT.}
\label{tab:3}
\vspace{-3mm}
\end{table}

%% file: tables/appendixA.tex
\begin{table}[h!]
\centering
\renewcommand{\tabcolsep}{2.0mm}
\begin{tabular}{cccc}
\toprule
\multicolumn{4}{c}{COCO Panoptic Classes} \\
\midrule
person & bicycle & car & motorcycle \\
airplane & bus & train & truck \\
boat & traffic light & fire hydrant & stop sign \\
parking meter & bench & bird & cat \\
dog & horse & sheep & cow \\
elephant & bear & zebra & giraffe \\
backpack & umbrella & handbag & tie \\
suitcase & frisbee & skis & snowboard \\
sports ball & kite & baseball bat & baseball glove \\
skateboard & surfboard & tennis racket & bottle \\
wine glass & cup & fork & knife \\
spoon & bowl & banana & apple \\
sandwich & orange & broccoli & carrot \\
hot dog & pizza & donut & cake \\
chair & couch & potted plant & bed \\
dining table & toilet & tv & laptop \\
mouse & remote & keyboard & cell phone \\
microwave & oven & toaster & sink \\
refrigerator & book & clock & vase \\
scissors & teddy bear & hair drier & toothbrush \\
banner$^*$ & blanket$^*$ & bridge$^*$ & cardboard$^*$ \\
counter$^*$ & curtain$^*$ & door$^*$ & floor-wood$^*$ \\
flower$^*$ & fruit$^*$ & gravel$^*$ & house$^*$ \\
light$^*$ & mirror$^*$ & net$^*$ & pillow$^*$ \\
platform$^*$ & playingfield$^*$ & railroad$^*$ & river$^*$ \\
road$^*$ & roof$^*$ & sand$^*$ & sea$^*$ \\
shelf$^*$ & snow$^*$ & stairs$^*$ & tent$^*$ \\
towel$^*$ & wall-brick$^*$ & wall-stone$^*$ & wall-tile$^*$ \\
wall-wood$^*$ & water$^*$ & window-blind$^*$ & window$^*$ \\
tree$^*$ & fence$^*$ & ceiling$^*$ & sky$^*$ \\
cabinet$^*$ & table$^*$ & floor$^*$ & pavement$^*$ \\
mountain$^*$ & grass$^*$ & dirt$^*$ & paper$^*$ \\
food$^*$ & building$^*$ & rock$^*$ & wall$^*$ \\
rug$^*$ \\
\bottomrule
\end{tabular}
\caption*{$*$: Object class that is not classified as `thing' (countable) but `stuff' (uncountable)}
\label{app:1}
\vspace{-5mm}
\end{table}

%% file: tables/table_ref.tex
\begin{table}[h!]
\resizebox{\linewidth}{!}{
\renewcommand{\tabcolsep}{3mm}
\begin{tabular}{lcccccccc}
\toprule
                     & \multicolumn{3}{c}{RefCOCO} & \multicolumn{3}{c}{RefCOCO+} & \multicolumn{2}{c}{RefCOCOg} \\
\cmidrule(lr){2-4}\cmidrule(lr){5-7}\cmidrule(lr){8-9}
Model                & val     & testA   & testB   & val      & testA   & testB   & val           & test         \\
\midrule
Shikra-7B            & 87.01   & 90.61   & 80.24   & 81.60    & 87.36   & 72.12   & 82.27         & 82.19        \\
Ferret-7B            & 87.49   & 91.35   & 82.45   & 80.78    & 87.38   & 73.14   & 83.93         & 84.76        \\
VistaLLM-7B          & 88.10   & 91.50   & 83.00    & 82.90   & 89.80   & 74.80   & 83.60         & 84.40         \\
Qwen-VL-7B           & 89.36   & 92.26   & 85.34   & 83.12    & 88.25   & 77.21   & 85.58         & 85.48        \\
CogVLM-Grounding-17B & 92.76   & 94.75   & 88.99   & 88.68    & 92.91   & 83.39   & 89.75         & 90.79        \\
\cdashline{1-9}\noalign{\vskip 0.5ex}
CoLLaVO-7B           & 87.34   & 91.08   & 82.39   & 80.87    & 86.36   & 73.20   & 82.44         & 82.33       \\
\bottomrule
\end{tabular}
}
\caption{Comparing object grounding performances of Shikra~\citep{chen2023shikra}, Ferret~\citep{you2024ferret}, VistalLLM~\citep{pramanick2023jack}, Qwen-VL~\citep{bai2023qwen}, CogVLM-Grounding~\citep{wang2023cogvlm}, and CoLLaVO on several object grounding benchmarks: RefCOCO, RefCOCO+, and RefCOCOg~\citep{kazemzadeh-etal-2014-referitgame}. Even though CoLLaVO did not use object grounding dataset (RefCOCO) in training phase, CoLLaVO shows comparable zero-shot object grounding performances, compared with (no zero-shot) other models specifically targeting object grounding task trained with RefCOCO grounding dataset.}
\end{table}

%% file: tables/appendixD.tex
\begin{table}[h!]   
\centering
\renewcommand{\tabcolsep}{2.0mm}
\resizebox{\linewidth}{!}{
\begin{tabular}{crccccccc} 
\toprule
Types & Metrics & LLaVA & MiniGPT-4 & MM-GPT & mPLUG-Owl & InstructBLIP & Shikra & CoLLaVO \\
\midrule

\multirow{4}{*}{\rotatebox[origin=c]{90}{Adversarial}}  & Accuracy  &49.7 &65.2 &50.0 &50.7 &72.1 &83.1 &\textbf{86.8} \\
\cdashline{2-9}\noalign{\vskip 0.5ex}
                                                        & Precision &49.6 &61.2 &50.0 &50.3 &65.1 &85.6 &\textbf{96.3} \\
\cdashline{2-9}\noalign{\vskip 0.5ex}
                                                        & Recall    &99.1 &82.9 &\textbf{100.} &99.3 &95.1 &79.6 &76.5 \\
\cdashline{2-9}\noalign{\vskip 0.5ex}
                                                        & F1-Score  &66.3 &70.4 &66.7 &66.8 &77.3 &82.5 &\textbf{85.2}\\
\midrule

\multirow{4}{*}{\rotatebox[origin=c]{90}{Random}}       & Accuracy  &50.4 &49.7 &50.1 &54.0 &\textbf{88.6} &86.9 &87.4 \\
\cdashline{2-9}\noalign{\vskip 0.5ex}
                                                        & Precision &50.2 &78.2 &50.1 &52.1 &84.1 &94.4 &\textbf{98.1} \\
\cdashline{2-9}\noalign{\vskip 0.5ex}
                                                        & Recall    &99.1 &82.2 &\textbf{100.} &99.6 &95.1 &79.3 &76.2 \\
\cdashline{2-9}\noalign{\vskip 0.5ex}
                                                        & F1-Score  &66.6 &80.1 &66.7 &68.4 &\textbf{89.3} &86.2 &85.8\\
\midrule

\multirow{4}{*}{\rotatebox[origin=c]{90}{Popular}}      & Accuracy  &49.9 &69.7 &50.0 &50.9 &82.8 &84.0 &\textbf{87.6} \\
\cdashline{2-9}\noalign{\vskip 0.5ex}
                                                        & Precision &49.9 &65.9 &50.0 &50.5 &76.3 &87.6 &\textbf{98.0} \\
\cdashline{2-9}\noalign{\vskip 0.5ex}
                                                        & Recall    &99.3 &81.9 &\textbf{100.} &99.4 &95.1 &79.2 &76.8 \\
\cdashline{2-9}\noalign{\vskip 0.5ex}
                                                        & F1-Score  &66.4 &73.0 &66.7 &66.9 &84.7 &83.2 &\textbf{86.1}\\
\bottomrule

\end{tabular}
}
\caption{Measuring four metrics: Accuracy, Precision, Recall, F1-score on three types of question answering to evaluate hallucination of vision language models: Adversarial, Random, and Popular in POPE.}
\end{table}